\documentclass[conference]{IEEEtran}
\IEEEoverridecommandlockouts
\usepackage{cite}
\usepackage{amsmath,amssymb,amsfonts}
\usepackage{algorithmic}
\usepackage{graphicx}
\usepackage{textcomp}
\usepackage{xcolor}
\usepackage{algorithm2e}
\usepackage{tabularx}
\usepackage{stmaryrd}

\usepackage{hyperref}

\RestyleAlgo{ruled}

\newtheorem{theorem}{\bf Theorem}[section]

\newtheorem{remark}[theorem]{\bf Remark}

\def\BibTeX{{\rm B\kern-.05em{\sc i\kern-.025em b}\kern-.08em
    T\kern-.1667em\lower.7ex\hbox{E}\kern-.125emX}}
\begin{document}

\title{High Performance Computing Applied to Logistic Regression: A CPU
and GPU Implementation Comparison
}

\author{\IEEEauthorblockN{MOHAMED MOUHAJIR}
\IEEEauthorblockA{\textit{ENSIAS, Mohammed V University} \\
 Rabat, Morocco,\\
mohamed\_mouhajir@um5.ac.ma}
\and
\IEEEauthorblockN{ MOHAMMED NECHBA}
\IEEEauthorblockA{\textit{ENSIAS, Mohammed V University} \\
 Rabat, Morocco,\\
mohammed\_nechba@um5.ac.ma}
\and
\IEEEauthorblockN{YASSINE SEDJARI}
\IEEEauthorblockA{\textit{ ENSIAS, Mohammed V University} \\
Rabat, Morocco,\\
yassine\_sedjari@um5.ac.ma}

}

\maketitle

\begin{abstract}
We present a versatile GPU-based parallel version of Logistic Regression (LR), aiming to address the increasing demand for faster algorithms in binary classification due to large data sets. Our implementation is a direct translation of the parallel Gradient Descent Logistic Regression algorithm proposed by X. Zou et al. \cite{GD_logistic}. Our experiments demonstrate that our GPU-based LR outperforms existing CPU-based implementations in terms of execution time while maintaining comparable f1 score. The significant acceleration of processing large datasets makes our method particularly advantageous for real-time prediction applications like image recognition, spam detection, and fraud detection. Our algorithm is implemented in a ready-to-use Python library available at : \url{https://github.com/NechbaMohammed/SwiftLogisticReg}.
\end{abstract}


\section{Introduction}
The advent of machine learning algorithms has brought about significant changes in the field of data analysis and has become a critical component of many industries. One particular application of machine learning is binary classification, which has gained widespread use in various fields, such as image recognition, spam detection, and fraud detection. Logistic regression (LR) is one of the most widely employed algorithms for binary classification and has proven to be highly effective in real-world scenarios. However, the exponential increase in dataset sizes has led to a demand for faster algorithms to process this data.

To address this need, parallel computing is one possible solution, which can enable faster processing of large datasets. High-Performance Computing (HPC) has recognized the need for parallel computing using hardware accelerators such as Graphics Processing Units (GPUs), which have become an integral part of mainstream computing systems according to researchers such as J. D. Owens et al. \cite{owens_gpu_computing}. In the field of machine learning, the use of HPC has increased in recent years due to the tremendous amount of data used for learning algorithms, which not only improves prediction accuracy but also increases the computational burden, as argued by A. Kerestely \cite{kerestely_2020_hpc_in_ML}.

Although LR is one of the most commonly used machine learning algorithms for binary classification, researchers have been exploring the use of GPU parallelization to speed up the process. Early approaches, such as those used by H. Peng et al. \cite{Peng_2013_evaluating}, were based on CPU-based methods that used big data frameworks such as Spark and Hadoop. Z. Wu et al. \cite{wu_2015_hyperspectral_images} developed the first GPU-based LR implementation in 2015, which was specific to solving the problem of hyperspectral image classification and lacked generality.

In 2016, P. Peretti \& F. Amenta \cite{peretti2016breast} claimed to have implemented the parallel version of LR to predict breast cancer, but did not provide details about their algorithm, raising concerns about the compatibility of their GPU-based parallel implementation with the original version of LR.

This paper aims to fill the literature gap by presenting a GPU-based parallel version of LR that is a pure translation of the parallel Gradient Descent Logistic Regression algorithm described by X. Zou et al. \cite{GD_logistic}. Our implementation is designed to be versatile and can be applied to a wide range of domains without specific constraints. The source code of our implementation is available at \url{https://github.com/NechbaMohammed/SwiftLogisticReg}, allowing readers to explore and use the algorithm in their own work.

\section{Logistic Regression Preliminaries: Algorithmic pseudo code and flow chart}

\subsection{Fundamentals of Logistic Regression}
Logistic Regression is a commonly used machine learning algorithm for binary classification. Its aim, as explained by T. Hastie et al. \cite{Hastie_LR_fundamentals}, is to estimate the probability of a binary outcome based on a given set of input features. The logistic function, also known as the sigmoid function, is utilized to map the input features to a probability value between 0 and 1. The model parameters are optimized by maximum likelihood estimation to minimize the difference between the predicted probabilities and the actual labels. The logistic regression model can be represented mathematically as follows:

Let $x$ be an input feature vector of n dimensions and let $y$ be a binary output variable. The logistic regression model assumes that the probability of y being $1$ given the input $x$ is given by:

$$p(y=1|x) = \frac{1}{1+\exp(-w^Tx-b)}$$

where $w$ is an n-dimensional weight vector and $b$ is the bias term. Weights and bias are learned from training data by maximizing the likelihood function.

$$L(w,b) = \prod_{i=1}^{m} p(y^{(i)}|x^{(i)};w,b)$$

where $y^{(i)}$ and $x^{(i)}$ are the i-th training label and the vector of characteristics, respectively. Parameters $w$ and $b$ are estimated by maximizing the logarithmic likelihood function:

\begin{align*}
\ell(w, b)= & \sum_{i=1}^m y^{(i)} \log \left(p\left(y^{(i)}=1 \mid x^{(i)} ; w, b\right)\right) \\
& +\left(1-y^{(i)}\right) \log \left(1-p\left(y^{(i)}=1 \mid x^{(i)} ; w, b\right)\right)
\end{align*}

The optimization problem can be solved using gradient descent, Newton's method, or other optimization algorithms.
\subsection{Sequential Logistic Regression}
The following algorithm outlines the Gradient Descent-based Logistic Regression method, as described by X. Zou et al. \cite{GD_logistic}:

\begin{algorithm}
\caption{Sequential version of Logistic Regression}\label{alg:Sequential_Logistic_Regression}
\textbf{INPUT}
 \begin{itemize}
     \item $ X \quad$ :  Matrix of shape m$\times$n (the input matrix)
    \item $Y \quad$ : Row vector of shape 1$\times$m (the targets)
    \item $\alpha \quad$ : Real number (learning rate)
    \item $\epsilon \quad$ : Real number (the tolerance)
    
 \end{itemize}   

\KwResult{ $W\quad$ :  Row vector of shape 1$\times$n ( weights)}

\While{ $\left(\left\|\nabla Loss\right\|> \epsilon\right)$}{
$y_{pred} = sigmoid(W*X^T)$\\
        $ error = Y - y_{pred}$\\
        $\nabla Loss = \frac{ 1}{m}  (error*X) $\\
        $W = W - \alpha * \nabla Loss$\\
}
\textbf{Return} $\quad W$
\end{algorithm}

Where  $sigmoid(.)$ is the following function  :\begin{equation}
sigmoid(x) = \frac{1}{1 + e^{-x}}, \quad x \in \mathbb{R}
\end{equation}

 The flowchart of this algorithm can be seen as:

\begin{figure}[!h]
 \centering
    \includegraphics[scale=0.7]{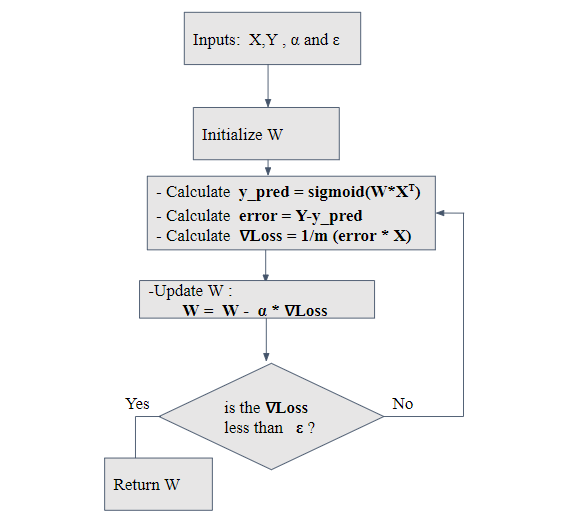}
    \caption{Flowchart for the sequential version of Logistic Regression }
    \label{fig:my_label}
\end{figure}  
\newpage
\section{Parallel Logistic Regression}

\subsection{Description of our approach}
One can implement parallel logistic regression by adopting well-established High Performance Computing strategies for Machine Learning, as illustrated in the literature, such as in \cite{Data_parallel_algorithms}, \cite{Principles_of_parallel_and_distributed_computing}, and \cite{Strategies_and_principles_of_distributed_machine_learning}. These methods offer multiple ways to achieve parallelism, which include:
\begin{itemize}
    \item \textbf{Data parallelism} : Dividing the data set into subsets and distributing them to different processors to perform the calculation in parallel.
    \item \textbf{model parallelism} : Splitting the model into sub-parts and distributing them to different processors to perform the calculation in parallel.
    \item \textbf{hybrid parallelism} : Combining data parallelism and model parallelism to perform the calculation in parallel.
\end{itemize}

In our case, we used \textbf{model parallelism}, which means we subdivided our algorithm into sub-parts, each of which is executed in parallel.

\begin{remark}
It is important to note that the parallel implementation of logistic regression is only beneficial when the data set is large enough to justify the overhead of parallelization, otherwise the sequential version is often more efficient.\\
\end{remark}

\subsection{Building blocks algorithms}
We have implemented our parallel logistic regression algorithm by decomposing each operation into a parallel version that can be executed on a Graphic Processing Unit (GPU). This approach has led to the creation of a set of algorithms that act as fundamental building blocks for our parallel logistic regression algorithm:
\subsubsection{Foundation Algorithms for Vector-Matrix Multiplication}

The algorithms \ref{alg:vector_matrix_mul} and \ref{alg:matrix_col_sum} serve as the foundation for carrying out the vector-matrix multiplication later on by the algorithm   \ref{alg:Paralle_version_of_Logistic_Regression} using a column-wise sum operation :

\begin{algorithm}[!h]
\caption{vector\_matrix\_mul}\label{alg:vector_matrix_mul}
\textbf{INPUT}
 \begin{itemize}
    \item $ M \quad$ : Matrix  of shape m$\times$n.
    \item $ v \quad$ : Row vector  of shape 1$\times$m.
 \end{itemize}   

\KwResult{ Modified version of M }

\For{$  i \in  \llbracket 1, m \rrbracket   \text{ in parallel }$}{
   \For { $j   \in \llbracket 1, n \rrbracket  $  } {
       $M[i][j] \gets v[0][i] * M[i][j]$ 
    } 
}
\end{algorithm}

\begin{algorithm}[!h]
\caption{matrix\_col\_sum}
\label{alg:matrix_col_sum}
\textbf{INPUT}
 \begin{itemize}
    \item $ M\quad$ : Matrix of shape m$\times$n .

    \item $ res \quad$ : Row vector  of shape 1$\times$n .
    
 \end{itemize}   

\KwResult{Modified version of $res$ }

\For{$  j \in \llbracket 1, n \rrbracket  \text{ in parallel }$}{
       $ res[0][j] \gets 0$\\
   \For { $i   \in \llbracket 1, m \rrbracket  $  } {
       $res[0][j] \gets res[0][j]+  M[i][j]$
    } 
}
\end{algorithm}

\subsubsection{ Parallel Computations for Subtraction, Norm, and Sigmoid}

Our parallel logistic regression algorithm \ref{alg:Paralle_version_of_Logistic_Regression} utilizes a parallel subtraction operation, as outlined in \ref{alg:Subtract}, along with the norm operator specified in \ref{alg:norm2}. 

\begin{algorithm}[!h]
\caption{norm2}\label{alg:norm2}
\textbf{ENTREE}
 \begin{itemize}
    \item $ v \quad$ : Row vector of shape 1$\times$n
    \item $v\_norm \quad$: Real Number
 \end{itemize}   

\KwResult{ $ v\_norm \quad$ : modified version of $v\_norm$ containing the norm of $v$}

\For{$  i \in \llbracket 1, n \rrbracket  \text{ in parallel }$}{
$v\_norm  \gets v\_norm+v[0][i]^2$

}
\end{algorithm}
\,\\
\begin{algorithm}[!h]
\caption{Substract}
\label{alg:Subtract}
\textbf{INPUT}
 \begin{itemize}

    \item $ vec1\quad$ :  Row vector of shape 1$\times$m
    \item $ res2 \quad$ : Row vector of shape 1$\times$m
 \end{itemize}   

\KwResult{  Modified version of  res2 }

\For{$  i \in \llbracket 1, m \rrbracket  \text{ in parallel }$}{
  $res2[0][i] \gets res2[0][i] - vec1[0][i]$
}
\end{algorithm}

In addition, a crucial aspect of the parallel logistic regression algorithm \ref{alg:Paralle_version_of_Logistic_Regression} is the computation of the sigmoid function, which will be carried out in parallel using algorithm \ref{alg:hypothesis}.

\begin{algorithm}[!h]
\caption{sigmoid}
\label{alg:hypothesis}
\textbf{INPUT}
 \begin{itemize}
     \item $ res \quad$ :Row vector of shape 1$\times$m
    \item  $ y_{pred} \quad$ :  Row vector of shape 1$\times$m
 \end{itemize}   

\KwResult{ $ y_{pred} \quad$ : modified version of $y_{pred}$}
\,\\
\For{$  i \in \llbracket 1, m \rrbracket  \text{ in parallel }$}{
$y_{pred}[0][i]\gets \dfrac{1}{(1+\exp(-res[0][i]))} $

}
\end{algorithm}

\newpage
\subsection{Parallel Logistic regression and flowchart }

\begin{figure}
\centering
    \includegraphics[scale=0.34]{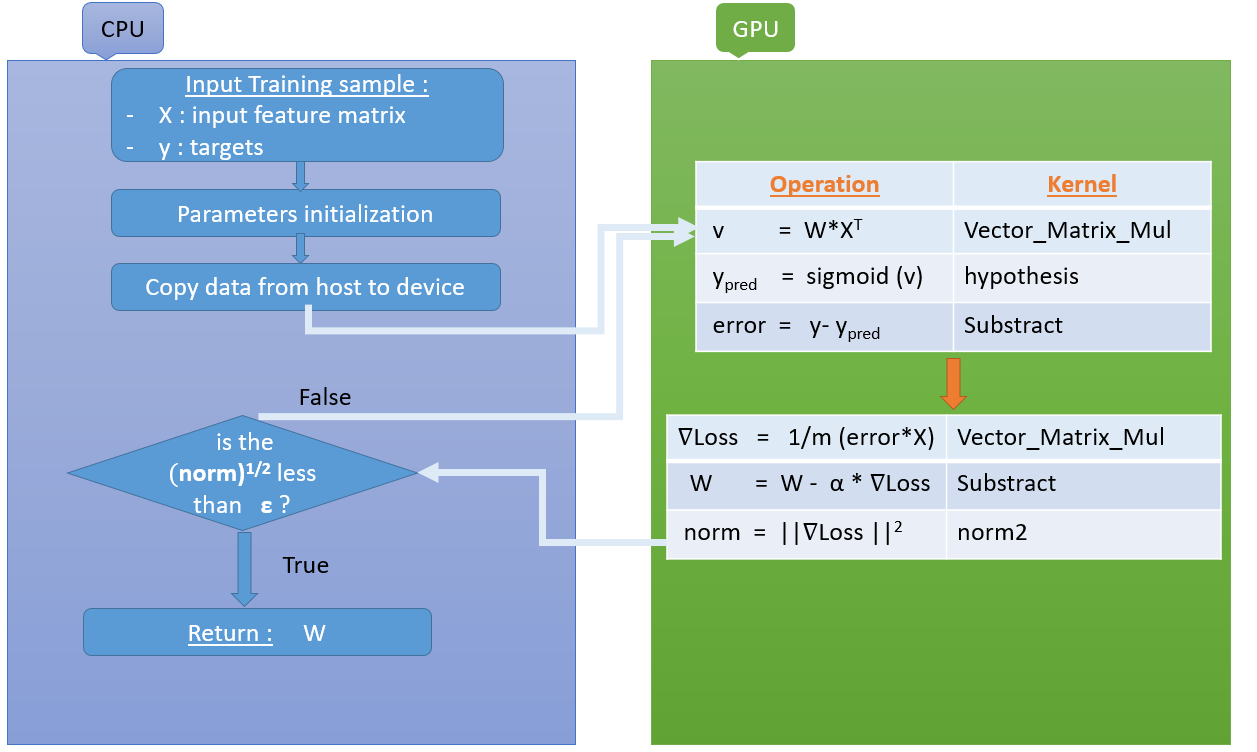}
    \caption{Flowchart for the Parallel Logistic Regression }
    \label{fig:Flowchat_for_the Parallel_Logistic_Regression}
\end{figure}

Once all the crucial components of our approach have been outlined, we will proceed to introduce our parallel logistic regression algorithm \ref{alg:Paralle_version_of_Logistic_Regression} and its flowchart \ref{fig:Flowchat_for_the Parallel_Logistic_Regression} in a formal manner :

\begin{algorithm}
\caption{ Parallel  Logistic Regression}
\label{alg:Paralle_version_of_Logistic_Regression}
\textbf{ENTREE}
 \begin{itemize}

    \item $ X \quad$ :  Matrix of shape m$\times$n
    \item $ Y \quad$ : Row vector of shape 1$\times$m
    \item $w \quad$: Row vector of shape 1$\times$n
    \item $\alpha \quad$: Real  number ( learning rate)
    \item $\epsilon \quad$: Real number ( the tolerance )
 \end{itemize}   

\KwResult{$w^*$ }
$\nabla Loss \gets w$\\
\While{true}{
$res \gets$ Row vector of shape 1$\times$m filled with 0\\
 vector\_matrix\_mul($w,X.T$)\\
 matrix\_col\_sum($X,res$)\\
 sigmoid($res$)\\
Substract( $Y,res$)\\
vector\_matrix\_mul($res,X$)\\
 matrix\_col\_sum($X,\nabla Loss$)\\
Substract( $\alpha*\nabla Loss,w$)\\
grad\_norm $\gets 0$\\
 norm2($\nabla Loss,grad\_norm$)\\
 \,\\
 \If { $\sqrt{(grad\_norm)}  \leq \epsilon$}{
 \,\\
 $w^*\gets w$
 \,\\
\textbf{Return}  $w^*$
 }    
}
\end{algorithm}

\newpage

\section{Experiments}
In this section, we evaluate the effectiveness of our proposed parallel Logistic Regression algorithm (Algorithm \ref{alg:Paralle_version_of_Logistic_Regression}) against two other approaches: the Sickit-learn implementation of Logistic Regression \cite{Scikit_learn}, and a sequential Gradient Descent-based logistic regression method, as described in \cite{GD_logistic} and implemented in Algorithm \ref{alg:Sequential_Logistic_Regression}. A comparative analysis is presented to highlight the performance of each approach.

\subsection{Experimental setting}
\subsubsection{Hardware}
We utilized the standard CPU and GPU provided by Google Colab. The specifications of the machine can be found in Table \ref{hardware}.

\begin{table}[!h]
    \centering

\caption{Hardware  specifications \label{hardware}}
    \begin{tabularx}{0.5\textwidth} { 
  | >{\raggedright\arraybackslash}X 
  | >{\centering\arraybackslash}X
  | >{\raggedleft\arraybackslash}X | }
 \hline
Resource & Details  \\
 \hline
  CPU  & Intel(R) Xeon(R) CPU @ 2.20GHz.  \\
\hline
GPU  &  NVIDIA Tesla K80\\
\hline
\end{tabularx}
\end{table}

\subsubsection{Dataset}
We utilized the publicly available HIGGS dataset in our experiments. The dataset contains simulated measurements of high-energy collisions of protons produced by the Large Hadron Collider (LHC) at CERN and consists of 11 million events. Each event is represented by 28 features, which include 21 low-level kinematic properties, 2 high-level features derived from the low-level features, and 5 expert features encoding the raw detector signals. The primary objective of the dataset is to differentiate between a signal process that produces Higgs bosons and a background process that does not. The HIGGS dataset was originally introduced in a research paper titled "Discovering the Higgs boson in the noise" by Baldi et al. (Nature Communications, 2014) \cite{HIGGS_Dataset}.

\newpage
\subsection{Results}

\begin{table}[!h]
    \centering
\caption{for $\alpha=0.1$ and $\epsilon =0.01$\label{tab_1}}
    \begin{tabularx}{0.49\textwidth} { 
  | >{\centering\arraybackslash}X
  | >{\centering\arraybackslash}X
  | >{\raggedleft\arraybackslash}X | }
 \hline
&  f$1$ score &execution time (seconds)  \\
 \hline
 sequential logistic regression   & 0.668  &73.383  \\
\hline
parallel logistic regression  &  \textbf{0.679 } &\textbf{14.779}\\
\hline
sklearn logistic regression   & 0.686 &19.378\\
\hline
\end{tabularx}
\end{table}

 \begin{table}[!h]
    \centering

\caption{for $\alpha =0.01$ and $\epsilon=0.008$
\label{tab_2}}
    \begin{tabularx}{0.52\textwidth} { 
  | >{\centering\arraybackslash}X
  | >{\centering\arraybackslash}X
  | >{\raggedleft\arraybackslash}X | }
 \hline
&  f$1$ score &execution time (seconds)  \\
 \hline
 sequential logistic regression  &0.668  &930.11  \\
\hline
parallel logistic regression  &  \textbf{0.692 }&\textbf{9.4}\\
\hline
sklearn logistic regression   & 0.686 &17.696\\
\hline
\end{tabularx}

\end{table}

From the experimental results, it is evident that our proposed GPU-based Parallel Logistic Regression (Parallel LR) algorithm \ref{alg:Paralle_version_of_Logistic_Regression} offers compelling advantages over traditional Sequential Logistic Regression (Sequential LR) \ref{alg:Sequential_Logistic_Regression} and the sklearn Logistic Regression (sklearn LR) implementations.

Regarding the f1 score, the proposed Parallel LR consistently exhibits competitive performance compared to both Sequential LR and sklearn LR. In Table \ref{tab_1}, for $\alpha=0.1$ and $\epsilon=0.01$, the Parallel LR achieves an f1 score of 0.679, outperforming Sequential LR's score of 0.668 and closely matching sklearn LR's score of 0.686. Similarly, in Table \ref{tab_2}, for $\alpha=0.01$ and $\epsilon=0.008$, the Parallel LR achieves the highest f1 score of 0.692, compared to Sequential LR's score of 0.668 and sklearn LR's score of 0.686.

More notably, the most significant advantage of the proposed Parallel LR is observed in terms of execution time. In both sets of experiments, our GPU-based Parallel LR remarkably reduces the execution time compared to both Sequential LR and sklearn LR. The reduced execution time can be seen in Table \ref{tab_1} and Table \ref{tab_2}, where the Parallel LR completes the computations in significantly less time (14.779 seconds and 9.4 seconds, respectively) compared to Sequential LR (73.383 seconds and 930.11 seconds, respectively) and sklearn LR (19.378 seconds and 17.696 seconds, respectively).

\subsection{interpretation of results}

The experimental evaluation of our GPU-based Parallel Logistic Regression (Parallel LR) algorithm against Sequential Logistic Regression (Sequential LR) and sklearn Logistic Regression (sklearn LR) reveals compelling insights.

Firstly, the competitive f1 scores achieved by the Parallel LR highlight its effectiveness in binary classification tasks. With scores comparable to or even slightly surpassing sklearn LR, our algorithm demonstrates its capability to capture underlying patterns in the data and make accurate predictions.

Secondly, the most significant advantage of the Parallel LR is its remarkable reduction in execution time. Leveraging the power of GPUs, the parallelization approach accelerates computations, yielding execution times that are considerably lower than those of both Sequential LR and sklearn LR. This efficiency gain is essential for real-time applications where rapid predictions are essential.

The combined advantages of competitive classification accuracy and superior execution time make our Parallel LR a compelling choice for diverse real-world applications. Its practical relevance, user-friendly integration, and adaptability to modern computing architectures position it as a promising solution for time-critical machine learning tasks.

In conclusion, the experimental results reaffirm the effectiveness and efficiency of our proposed GPU-based Parallel Logistic Regression algorithm, showcasing its potential to contribute significantly to the field of machine learning and real-world applications.

\section{Conclusion}
In the context of real-world machine learning applications, the size of the dataset often poses a significant challenge, with execution time being a crucial factor in selecting an appropriate learning algorithm. To overcome this challenge, high-performance computing techniques, especially parallelism, have emerged as a promising solution.

In this paper, we proposed a GPU-based parallel version of logistic regression using the gradient descent optimizer and conducted an experimental evaluation. Our results indicate that our algorithm outperforms the sequential and built-in sklearn algorithms in terms of execution time. However, there may be cases where our algorithm falls short in terms of f1 score due to the choice of hyperparameters.

Future work on this topic could include implementing logistic regression using other optimizers and incorporating commonly used techniques in machine learning, such as regularization, to avoid overfitting. In general, our study provides a foundation for further exploration of the potential of high-performance computing techniques in machine learning applications.

\bibliographystyle{plain}
\bibliography{IEEE}

\begin{thebibliography}{10}

\bibitem{HIGGS_Dataset}
Pierre Baldi, Peter Sadowski, and Daniel Whiteson.
\newblock Searching for exotic particles in high-energy physics with deep
  learning.
\newblock {\em Nature communications}, 5(1):4308, 2014.

\bibitem{Principles_of_parallel_and_distributed_computing}
Rajkumar Buyya, Christian Vecchiola, and S~Thamarai Selvi.
\newblock Principles of parallel and distributed computing.
\newblock {\em Mastering cloud computing}, pages 29--70, 2013.

\bibitem{Hastie_LR_fundamentals}
Trevor Hastie, Robert Tibshirani, Jerome~H Friedman, and Jerome~H Friedman.
\newblock {\em The elements of statistical learning: data mining, inference,
  and prediction}, volume~2.
\newblock Springer, 2009.

\bibitem{Data_parallel_algorithms}
W~Daniel Hillis and Guy~L Steele~Jr.
\newblock Data parallel algorithms.
\newblock {\em Communications of the ACM}, 29(12):1170--1183, 1986.

\bibitem{kerestely_2020_hpc_in_ML}
Arpad Kerestely.
\newblock High performance computing for machine learning.
\newblock {\em Bulletin of the Transilvania University of Brasov. Series III:
  Mathematics and Computer Science}, pages 705--714, 2020.

\bibitem{owens_gpu_computing}
John~D Owens, Mike Houston, David Luebke, Simon Green, John~E Stone, and
  James~C Phillips.
\newblock Gpu computing.
\newblock {\em Proceedings of the IEEE}, 96(5):879--899, 2008.

\bibitem{Scikit_learn}
Fabian Pedregosa, Ga{\"e}l Varoquaux, Alexandre Gramfort, Vincent Michel,
  Bertrand Thirion, Olivier Grisel, Mathieu Blondel, Peter Prettenhofer, Ron
  Weiss, Vincent Dubourg, et~al.
\newblock Scikit-learn: Machine learning in python.
\newblock {\em the Journal of machine Learning research}, 12:2825--2830, 2011.

\bibitem{Peng_2013_evaluating}
Haoruo Peng, Ding Liang, and Cyrus Choi.
\newblock Evaluating parallel logistic regression models.
\newblock In {\em 2013 IEEE International Conference on Big Data}, pages
  119--126. IEEE, 2013.

\bibitem{peretti2016breast}
A~Peretti and F~Amenta.
\newblock Breast cancer prediction by logistic regression with cuda parallel
  programming support.
\newblock {\em Breast Can Curr Res}, 1(111):2, 2016.

\bibitem{wu_2015_hyperspectral_images}
Zebin Wu, Qicong Wang, Antonio Plaza, Jun Li, Le~Sun, and Zhihui Wei.
\newblock Real-time implementation of the sparse multinomial logistic
  regression for hyperspectral image classification on gpus.
\newblock {\em IEEE Geoscience and Remote Sensing Letters}, 12(7):1456--1460,
  2015.

\bibitem{Strategies_and_principles_of_distributed_machine_learning}
Eric~P Xing, Qirong Ho, Pengtao Xie, and Dai Wei.
\newblock Strategies and principles of distributed machine learning on big
  data.
\newblock {\em Engineering}, 2(2):179--195, 2016.

\bibitem{GD_logistic}
Xiaonan Zou, Yong Hu, Zhewen Tian, and Kaiyuan Shen.
\newblock Logistic regression model optimization and case analysis.
\newblock In {\em 2019 IEEE 7th international conference on computer science
  and network technology (ICCSNT)}, pages 135--139. IEEE, 2019.

\end{thebibliography}

\end{document}